\RequirePackage{amsthm}
\documentclass[sn-mathphys-num,iicol]{sn-jnl}% Math and Physical 
\usepackage{graphicx}%
\usepackage{float}
\usepackage{multirow}%
\usepackage{amsmath,amssymb,amsfonts}%
\usepackage{mathrsfs}%
\usepackage[title]{appendix}%
\usepackage{xcolor}%
\usepackage{textcomp}%
\usepackage{manyfoot}%
\usepackage{booktabs}%
\usepackage{algorithm}%
\usepackage{algorithmicx}%
\usepackage{algpseudocode}%
\usepackage{listings}%
\usepackage{stfloats}
\usepackage{graphicx}
\usepackage{array}
\usepackage[table]{xcolor}
\usepackage{booktabs}
\usepackage{multirow}
\newcommand{\orcidlink}[1]{%
    \href{https://orcid.org/#1}{\includegraphics[height=1.5ex]{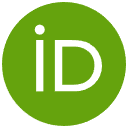}}%
}
\definecolor{lightred}{RGB}{255,0,0} 
\definecolor{lightblue}{RGB}{0,0,255}
\definecolor{lightgray}{RGB}{245,245,245}
\definecolor{lightpurple}{RGB}{245, 235, 255}
\theoremstyle{thmstyleone}%
\theoremstyle{thmstyletwo}%

\theoremstyle{thmstylethree}%

\begin{document}

\title[Article Title]{SCEESR: Semantic-Control Edge Enhancement for Diffusion-Based Super-Resolution}

%%=============================================================%%
%% GivenName	-> \fnm{Joergen W.}
%% Particle	-> \spfx{van der} -> surname prefix
%% FamilyName	-> \sur{Ploeg}
%% Suffix	-> \sfx{IV}
%% \author*[1,2]{\fnm{Joergen W.} \spfx{van der} \sur{Ploeg} 
%%  \sfx{IV}}\email{iauthor@gmail.com}
%%=============================================================%%

\author*[1]{\fnm{Yun Kai} \sur{Zhuang} \orcidlink{0009-0007-5476-4400}}\email{zhuangyk2023@shanghaitech.edu.cn}

% \author[2,3]{\fnm{Second} \sur{Author}}\email{iiauthor@gmail.com}
% \equalcont{These authors contributed equally to this work.}

% \author[1,2]{\fnm{Third} \sur{Author}}\email{iiiauthor@gmail.com}
% \equalcont{These authors contributed equally to this work.}

\affil*[1]{\orgdiv{School of Information Science and Technology}, \orgname{ShanghaiTech University}, \orgaddress{\street{393 Middle Huaxia Road}, \city{Shanghai}, \postcode{201210}, \country{China}}}

% \affil[2]{\orgdiv{Department}, \orgname{Organization}, \orgaddress{\street{Street}, \city{City}, \postcode{10587}, \state{State}, \country{Country}}}

% \affil[3]{\orgdiv{Department}, \orgname{Organization}, \orgaddress{\street{Street}, \city{City}, \postcode{610101}, \state{State}, \country{Country}}}

%%==================================%%
%% Sample for unstructured abstract %%
%%==================================%%

\abstract{Real-world image super-resolution (Real-ISR) must handle complex degradations and inherent reconstruction ambiguities. While generative models have improved perceptual quality, a key trade-off remains with computational cost. One-step diffusion models offer speed but often produce structural inaccuracies due to distillation artifacts. To address this, we propose a novel SR framework that enhances a one-step diffusion model using a ControlNet mechanism for semantic edge guidance. This integrates edge information to provide dynamic structural control during single-pass inference. We also introduce a hybrid loss combining L2, LPIPS, and an edge-aware AME loss to optimize for pixel accuracy, perceptual quality, and geometric precision. Experiments show our method effectively improves structural integrity and realism while maintaining the efficiency of one-step generation, achieving a superior balance between output quality and inference speed. The results of test datasets will be published at {\href{https://drive.google.com/drive/folders/1amddXQ5orIyjbxHgGpzqFHZ6KTolinJF?usp=drive\_link}{here}} and the related code will be published at {\href{https://github.com/ARBEZ-ZEBRA/SCEESR}{here}}.}


%%\pacs[JEL Classification]{D8, H51}

%%\pacs[MSC Classification]{35A01, 65L10, 65L12, 65L20, 65L70}

\maketitle

\section{Introduction}\label{sec1}

Real-world image super-resolution (Real-ISR) is a challenging and rapidly evolving research area aimed at reconstructing high-quality (HQ) images from their low-quality (LQ) counterparts. These LQ inputs may be captured directly in real-world conditions or synthesized using simulated degradation models that often incorporate complex and unpredictable distortions such as noise, blur, and non-ideal downsampling. A fundamental difficulty in real-world SR stems from the inherent ambiguity in the degradation process and the irreversible loss of high-frequency information, making it difficult to recover realistic texture details.

Initially, Convolutional Neural Networks (CNNs) with substantial depth were established as the backbone for learning the complex LQ-to-HQ mapping, primarily optimized with L1/L2 losses. To escape the perceptual plateau of these deterministic models, Generative Adversarial Networks (GANs) introduced a paradigm shift. The core principle involves a generator that produces the SR image and a discriminator that distinguishes it from real HQ images, which directly encourages the generation of highly realistic and textured outputs. More recently, Diffusion Models (DMs) have emerged as a powerful alternative. Their principle is based on a two-stage process: a forward pass that progressively adds noise to a ground-truth image until it becomes pure noise, and a reverse pass where a neural network is trained to iteratively denoise a random vector, conditioned on the LQ image, to reconstruct the HQ output. This iterative refinement process allows DMs to capture a rich distribution of image details, often yielding superior diversity and fidelity in generated textures compared to GANs.

To address the significant computational bottleneck of iterative sampling in diffusion models, researchers have developed one-step generation. The core principle involves distilling the complex multi-step denoising process of a pre-trained diffusion model into a single, much larger neural network pass. Despite these advancements, one-step diffusion models still exhibit certain limitations. A primary shortcoming is their tendency to occasionally produce outputs with compromised structural integrity or subtle artifacts, as the single-pass generation may fail to fully capture complex, high-frequency details and precise edge information that are critical for perceptual realism. This is partly because the distillation process, which compresses the multi-step denoising pipeline into one step, can lead to a loss of granular control over structural elements. 

To overcome these limitations, our work introduces a key innovation by integrating a ControlNet mechanism\citep{zhang2023adding} into a one-step diffusion model. This enhancement allows the model to be dynamically guided by semantic input conditions, specifically by adjusting the generative weights based on different types of edge maps, such as Canny\citep{canny2009computational} and HED (Holistically-Nested Edge Detection)\citep{xie2015holistically}. The core idea is to empower the model to emphasize critical edge information during the single-pass synthesis process. The ControlNet acts as a conditional adapter, processing the input edge maps and injecting this structural guidance into the diffusion model's sampling process. This ensures that the generated high-resolution images not only possess realistic textures but also maintain accurate and sharp structural outlines.

Furthermore, to explicitly reinforce the learning of edge details, we augment the loss function by incorporating an edge-aware AME loss term. This term specifically penalizes discrepancies between the edge maps of the super-resolved output and the ground-truth high-resolution image. By combining this AME loss with traditional L2 and LPIPS\citep{zhang2018unreasonable} losses, our hybrid loss function simultaneously optimizes for pixel-level accuracy, overall structural similarity, and geometric precision along edges. As a result, our method achieves a superior balance, enhancing the perceptual quality and structural fidelity of the generated images while maintaining the computational efficiency of a one-step generation framework.

\section{Related Work}\label{sec2}

The pursuit of high-fidelity image super-resolution has led to significant methodological evolution, primarily driven by advancements in deep neural networks. This section reviews the key developments in loss functions, network architectures, and models that form the foundation for our work.

\subsection{Evolution of Loss Functions for Perceptual Quality} \label{subsec21}
Early deep learning-based SR models predominantly relied on pixel-wise regression losses, such as L1 or L2 (MSE) loss. While these losses are effective at minimizing the average pixel error and achieving high Peak Signal-to-Noise Ratio (PSNR), they inherently lead to perceptually unsatisfactory, over-smoothed reconstructions with a lack of high-frequency texture. To address this, the research community shifted towards perception-driven metrics. The Structural Similarity Index (SSIM) loss was introduced to better align with human vision by measuring local patterns of luminance and contrast. A more significant leap was made with the advent of perceptual losses, most notably the Learned Perceptual Image Patch Similarity (LPIPS)\citep{zhang2018unreasonable} metric, which leverages features from pre-trained deep networks to measure semantic similarity in a high-dimensional feature space, encouraging outputs that are perceptually convincing even if pixel-level accuracy is slightly compromised.

\subsection{From CNNs to Diffusion Models} \label{subsec22}
The architecture of SR models has evolved in tandem with their training objectives. Pioneering works employed deep Convolutional Neural Networks (CNNs)\citep{dong2015image}, which have difficulties in perceptual losses. 

A paradigm shift in super-resolution occurred with the introduction of Generative Adversarial Networks (GANs). Pioneering works such as SRGAN\citep{ledig2017photo} demonstrated that adversarial training could yield results with highly realistic textures and significantly improved perceptual quality. Subsequent models like BSRGAN\citep{zhang2021designing} and RealESRGAN\citep{wang2021real} further advanced real-world image super-resolution (Real-ISR) by handling more complex and diverse degradations. FeMaSR\citep{chen2022real} recovers realistic high-resolution images by matching distorted low-resolution features to pretrained HR priors in a compact feature space, avoiding GAN instability without needing explicit references. Despite these advancements, GAN-based methods face inherent limitations. The adversarial training process is often unstable, and the discriminator's ability to assess the quality of diverse natural image content remains limited. 

More recently, Diffusion Models (DMs) have emerged as a state-of-the-art generative framework. DMs excel at capturing a rich distribution of natural image details, often yielding outputs with superior diversity and fidelity compared to GANs.

\subsection{From Multi-Steps Diffusion Models to One-Step Diffusion Models} \label{subsec23}
Recent studies have increasingly explored multi-step diffusion models for image super-resolution (SR). For instance, ResShift\citep{yue2023resshift} trains a diffusion model from scratch on paired low-quality and high-quality (LQ-HQ) image data. StableSR\citep{wang2024exploiting} introduces a trainable encoder and uses the LQ image as a conditioning input to guide a pre-trained Stable Diffusion model. PASD\citep{khan2023pasd} incorporates an encoder for degradation removal and further proposes a pixel-aware cross-attention module to integrate both low-level and high-level image features into the diffusion process. SeeSR\citep{wu2024seesr} improves semantic robustness by leveraging degradation-aware tag-style prompts to steer the generative process. However, the multi-step sampling procedure in such models leads to increased computational cost and a higher risk of generating unrealistic or unfaithful image content. 

To mitigate these issues, researchers have begun developing diffusion-based SR methods with reduced sampling steps. SinSR\citep{wang2024sinsr} applies consistency-preserving distillation to accelerate the diffusion process originally introduced in ResShift\cite{yue2023resshift}. Meanwhile, OSEDiff\citep{wu2024one} adopts the LQ image as the direct input—bypassing random noise sampling—and utilizes variational score distillation (VSD)\citep{wang2023prolificdreamer, yin2024one} loss to distill the generative capability of a multi-step diffusion model into a one-step inference framework, thereby providing an efficient DM-based SR solution.

\subsection{The Role of Structural Guidance} \label{subsec24}
Incorporating explicit structural guidance is crucial for enhancing the geometric fidelity of generative models. This has led to the use of various edge detection techniques to guide the image generation process. The Sobel operator, for instance, is a classic method that approximates the image gradient to highlight regions of high spatial frequency. The Laplacian of Gaussian (LoG) filter, another foundational technique, identifies edges by locating zero-crossings after applying a Laplacian filter to a Gaussian-smoothed image. More advanced methods like the Canny detector\citep{canny2009computational} provide cleaner, well-localized edges by employing non-maximum suppression and hysteresis thresholding. Similarly, Holistically-nested Edge Detection (HED)\citep{xie2015holistically} is a deep learning-based approach that produces semantically rich edge maps by leveraging multi-scale features.

The advent of ControlNet\citep{zhang2023adding} has provided a powerful framework for adding spatial conditioning to large-scale text-to-image multi-steps diffusion models. A prominent application\citep{bevacqua2024enhancing} is the use of Canny edge detection to steer the generative process, ensuring that the output adheres to specific structural outlines.

Our work integrates a ControlNet-like guidance mechanism, conditioned on such edge information, into a one-step diffusion model. This adapter provides granular structural control, directly addressing the inaccuracies of one-step generation. Furthermore, we augment the training with an edge-aware loss to explicitly reinforce geometric detail learning, achieving a superior balance between perceptual quality and structural fidelity.

\section{Methodology} \label{sec3}
\begin{figure*}[htbp] 
    \centering
    \includegraphics[width=0.9\textwidth]{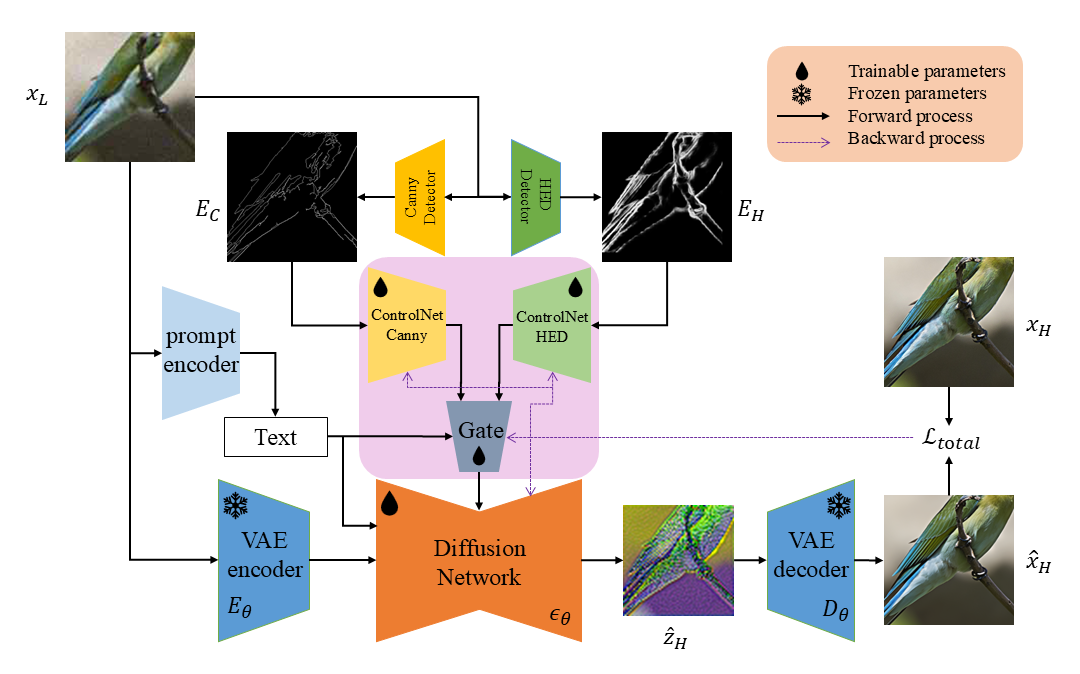}
    
    \caption{The training process of SCEESR. The LQ image is passed through a frozen encoder $E_\theta$, a LoRA finetuned diffusion network $\epsilon_\theta$ and a frozen decoder $D_\theta$. Text prompts, generated by a prompt encoder, are fed into both the diffusion network $\epsilon_\theta$ and a semantic control gate. Simultaneously, Canny edge maps and HED edge maps are extracted from the LQ image using a Canny detector and an HED detector, respectively. The outputs from the two ControlNets—corresponding to the edge maps—are fused in the semantic control gate under the guidance of the text prompts. The fused control information is then injected into each layer of the diffusion network $\epsilon_\theta$. During inference, only the diffusion network $\epsilon_\theta$, the two ControlNets, and the parameters of the semantic control gate are utilized.}
    
    \label{fig:train_process}
\end{figure*}

Our proposed framework is designed to enhance the structural fidelity of one-step diffusion models for real-world image super-resolution. The overall architecture, illustrated in the Figure~\ref{fig:train_process}, integrates a gated ControlNet mechanism and a hybrid loss to guide the generation process. The following subsections detail its core components. 

\subsection{One-Step Diffusion Model for Super-Resolution} \label{subsec31}
As discussed, existing diffusion-based Real-ISR methods typically require multiple timesteps to estimate the HQ image from random noise initialization with LQ images as conditional inputs. These approaches are computationally expensive and inherently introduce stochasticity. We propose a super-resolution network based on one-step diffusion for Real-ISR.

Our generator $G_\theta$ consists of three key components: a frozen encoder $E_\theta$, a fine-tuned diffusion network $\epsilon_\theta$, and a frozen decoder $D_\theta$. Let $E_\phi$, $\epsilon_\phi$ and $D_\phi$ denote the VAE encoder, latent diffusion network, and VAE decoder of a pretrained Stable Diffusion (SD) model, where $\phi$ represents the pretrained parameters. Following recent success of LoRA\citep{shah2024ziplora,stracke2024ctrloralter,jones2024customizing} in fine-tuning SD for downstream tasks, we employ LoRA to adapt the pretrained SD model for Real-ISR.

We maintain SD's original generation capacity by introducing trainable LoRA layers to the diffusion network $\epsilon_\phi$, transforming them into $\epsilon_\theta$ through fine-tuning with our training data. The encoder paremeters and decoder parameters remain fixed ($D_\theta = D_\phi$) to ensure consistency between the diffusion network output.

Recall that the diffusion process transforms input latent $z$ through $z_t = \alpha_t z + \beta_t \epsilon$, where $\alpha_t,\beta_t$ are timestep-dependent scalars for $t \in \{1,\dots,T\}$. Given a neural network that predicts noise in $z_t$ as $\hat{\epsilon}$, the denoised latent can be obtained as:
\begin{equation}
\hat{z}_0 = \frac{z_t - \beta_t \hat{\epsilon}}{\alpha_t}
\end{equation}
which should yield cleaner and more photorealistic results than $z_t$. The original SD performs text-conditioned generation by extracting text embeddings $c_y$ from text description $y$, enabling noise prediction as $\hat{\epsilon} = \epsilon_\theta(z_t; t, c_y)$.

We adapt this text-to-image denoising process for Real-ISR by formulating the LQ-to-HQ latent transformation $F_\theta$ as a text-conditioned image-to-image denoising process:
\begin{equation}
\hat{z}_H = \frac{z_L - \beta_T \epsilon_\theta(z_L; T, c_y)}{\alpha_T}
\end{equation}
where we perform single-step denoising on the LQ latent $z_L$ at the final timestep $T$ without adding noise. The text embeddings $c_y = \mathcal{Y}(x_L)$ are extracted from the LQ input $x_L$ using a prompt extractor\citep{zheng2024dape}. The complete LQ-to-HQ image synthesis is:

\begin{equation}
\hat{x}_H = D_\theta(F_\theta(E_\theta(x_L); \mathcal{Y}(x_L)))
\end{equation}

\subsection{Structural Guidance with Canny and HED Edge Detectors} \label{subsec32}
\begin{figure}[htbp] 
    \centering
    \includegraphics[width=0.45\textwidth]{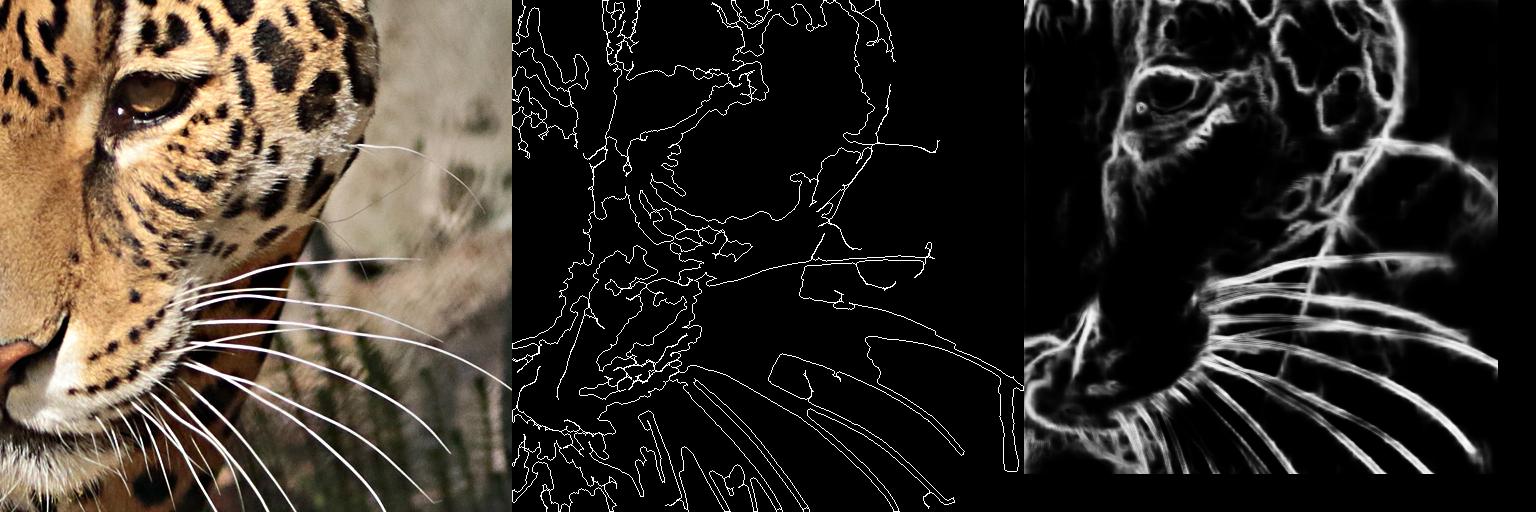}
    
    \caption{The figure presents a comparative visualization of edge detection results from the same input image (left). The middle image shows the output of the Canny detector\citep{canny2009computational} The right image displays results from the HED detector\citep{xie2015holistically}}
    
    \label{fig:edge_comparison}
\end{figure}

To compensate for the structural limitations of one-step generation, we employ explicit edge guidance. We extract edge maps from the upsampled LR image using two complementary detectors: Canny\citep{canny2009computational} and Holistically-Nested Edge Detection (HED)\citep{xie2015holistically}.

Canny Detector: The Canny operator is a classic multi-stage algorithm involving Gaussian filtering, gradient calculation, non-maximum suppression, and hysteresis thresholding. It is renowned for its strong response to prominent, sharp edges and is less sensitive to texture noise. This makes it highly effective for capturing clear structural boundaries and object contours.

HED Detector: HED is a deep learning-based approach that performs image-to-image prediction by leveraging holistically-nested networks with deep supervision. HED excels at capturing richer edge information, including semantically meaningful soft edges and fine-grained textures that the Canny detector might omit. However, it may also be more susceptible to producing thicker or noisier edge lines.

Performance Comparison and Rationale: The choice of these two detectors is strategic. As shown in the Figure~\ref{fig:edge_comparison}, the Canny detector provides precise geometric guidance for strong, unambiguous edges, ensuring structural integrity. In contrast, the HED detector provides comprehensive semantic guidance, capturing finer details and object-level boundaries that contribute to perceptual richness. By leveraging both, our model can access a more complete and multi-faceted representation of the image's structural information.

\subsection{Gated ControlNet with MLP-based Fusion} \label{subsec33}
To dynamically and adaptively fuse the complementary edge features from the Canny and HED detectors, we introduce a gated ControlNet mechanism driven by a Multi-Layer Perceptron (MLP). Instead of simply concatenating the edge maps, this design allows the model to learn a content-aware fusion strategy.

The process begins by extracting structural conditions from the low-resolution input. The Canny edge map $E_C$ and the HED edge map $E_H$ are first processed by their respective ControlNet adapters. These adapters, which are trainable copies of specific encoder blocks from the generator $G_{\theta}$, encode the edge maps into high-dimensional feature representations, denoted as $F_C = ControlNet_C(E_C)$ and $F_H = ControlNet_H(E_H)$.

The core innovation is a gating function implemented by a compact MLP, which acts as a semantic router. This MLP takes as input a globally pooled semantic feature vector $z_{sem}$ extracted from the prompt extractor. This vector $z_{sem}$ encapsulates the global semantic context of the input image. The MLP $g_\phi$ , parameterized by weights $\phi$, then maps this semantic vector to a set of fusion weights:
\begin{equation}
[\beta_C, \beta_H] = Softmax(g_{\phi}(z_{sem}))
\end{equation}
Here, $\beta_C$ and $\beta_H$ are scalar weights that control the global contribution of the Canny-guided and HED-guided features, respectively. The Softmax function ensures $\beta_C + \beta_H = 1$.

The final, fused edge guidance feature $F_{edge}$ that is injected into the Unet is computed as a weighted sum of the two ControlNet outputs: 
\begin{equation}
F_{edge} = \beta_C \cdot F_C + \beta_H \cdot F_H
\end{equation}

This gating mechanism enables a dynamic and semantically-informed interaction between the two structural guidance streams. Based on the input image's content, the MLP can learn to assign a higher weight $\beta_C$ to the Canny stream for precise geometric boundaries or a higher weight $\beta_H$ to the HED stream for richer semantic edges and textures. This adaptive fusion ensures that the most relevant structural guidance is emphasized during the single-pass synthesis, leading to outputs with enhanced structural integrity.

\subsection{Hybrid Loss Function} \label{subsec34}
To train the entire model end-to-end and explicitly enforce geometric accuracy, we employ a hybrid loss function $\mathcal{L}_{total}$ that combines pixel-level, perceptual, and edge-aware objectives: 
\begin{equation}
\mathcal{L}_{total} = \lambda_{L2} \mathcal{L}_{L2} + \lambda_{lpips} \mathcal{L}_{lpips} + \lambda_{AME} \mathcal{L}_{AME}
\end{equation}

Pixel Reconstruction Loss ($\mathcal{L}_2$): We use the L2 loss to ensure basic pixel-level fidelity between the generated image $\hat{x}_H$ and the ground-truth high-resolution image $x_H$.This term provides a stable training signal for the generator.

Perceptual Loss ($\mathcal{L}_{lpips}$): To improve the perceptual quality, we employ the LPIPS loss\citep{zhang2018unreasonable}, which measures the similarity in the feature space of a pre-trained network. This loss encourages the output to be semantically similar to the target.

Adaptive Multi-Detector Edge Loss ($\mathcal{L}_{AME}$) is based on the Entropy Weight Method. This loss function objectively evaluates the amount of information contributed by different edge detectors for each training batch and dynamically allocates weights, thereby achieving a more scientific and adaptive edge constraint. Our approach utilizes a set of four edge detectors to form a comprehensive edge-sensing system: the Sobel operator, which provides dense gradient magnitude information and is sensitive to subtle intensity changes; the Laplacian of Gaussian (LoG) operator, which offers better capture of edge scale characteristics; the Canny detector\citep{canny2009computational}, which produces fine, well-connected structural edges; and the Holistically-Nested Edge Detection (HED) detector\citep{xie2015holistically}, which extracts semantic-level edge information. For each detector $d$, we calculate its L1 loss and Structural Similarity (SSIM) loss components, defined as 
\begin{equation}
\mathcal{L}_{L1}^d = \|E_d(x_H)-E_d(\hat{x}_H)\|_1
\end{equation}
and 
\begin{equation}
\mathcal{L}_{SSIM}^d = 1 - SSIM(E_d(x_H), E_d(\hat{x}_H))
\end{equation}
, respectively.

The weighting process based on the Entropy Weight Method involves four key steps. First, for a batch of $N$ samples, a loss matrix $X\in R^{N\times 8}$ is constructed. Each column of this matrix corresponds to the values of a specific loss item across all samples in the batch. Second, each column of the loss matrix is normalized to the range [0, 1] using min-max standardization. Third, the information entropy for the j-th loss item is calculated as $e_{j}=-\frac{1}{\ln(N)}\sum_{i=1}^{N}p_{ij}\ln(p_{ij})$
, where $p_{ij} = \frac{r_{ij}}{\sum_{i=1}^Nr_{ij}}$ is the proportion value. Finally, the weight for each loss component is determined as 
\begin{equation}
\omega_j = \frac{1-e_j}{\sum_{k=1}^M(1-e_k)}
\end{equation}
. 

The final form of the proposed Adaptive Multi-Detector Edge Loss (AME) is defined based on the adaptively obtained weights from the Entropy Weight Method: 
\begin{equation}
\mathcal{L}_{AME} = \sum_{d\in D}(\omega_{L1}^d\cdot\mathcal{L}_{L1}^d+\omega_{SSIM}^d\cdot\mathcal{L}_{SSIM}^d)
\end{equation}
, where $\omega_{L1}^d$ and $\omega_{SSIM}^d$ are the weights derived from the Entropy Weight Method for the L1 loss and SSIM loss of detector d, respectively.

By jointly optimizing with this hybrid loss, our model is guided to produce images that are not only photorealistic in texture but also geometrically faithful to the ground-truth structure, thereby achieving a superior balance within an efficient one-step generation framework.

\section{Experiments} \label{sec4}
\subsection{Experiment Settings} \label{subsec41}
\subsubsection*{Implementation Details} \label{subsubsec411}
We train SCEESR with the AdamW optimizer at a learning rate of 5e-5 upon SD 2.1-base for the $\times$4 SR task. The rank of LoRA in the diffusion network is set to 4. We adopt the degradation-aware prompt extraction(DAPE)\citep{zheng2024dape} module in SeeSR\citep{wu2024seesr} to extract text prompts. The canny edge dector\citep{canny2009computational} applies Gaussian blur (7×7 kernel) and CLAHE with 2.0 cliplimit and uses the median pixel intensity and morphological closing (5×5 elliptical kernel). We use the deploy prototxt and caffemodel from Holistically-Nested Edge Detection\citep{xie2015holistically} to implement HED detector. The network structure of the gated function MLP consists of three layers and the output size is 2$\times$1. The batch size is 8 and the training patch size is 512$\times$512. The modules on 4 NVIDIA V100 GPUs undergo 8K training iterations with no inserting of the ControlNet firstly to avoid early interference and ensure complete convergence and undergo 8K training iterations with the ControlNet on the frozen UNet. The AME loss function is used in the second training part to enhance edges and the loss weights calculate through Entropy Weight Method based on the training datasets.

\subsubsection*{Training and Testing Datasets} \label{subsubsec412}
We train SCEESR on a combination of the LSDIR\citep{li2023lsdir} dataset and the first 10,000 face images from FFHQ\citep{karras2019style}. Low-quality (LQ) and high-quality (HQ) training pairs are synthesized using the degradation pipeline from Real-ESRGAN\citep{wang2021real}.
For evaluation, we compare SCEE-SR against several competing methods on the test set introduced by StableSR\citep{wang2024exploiting}, which comprises both synthetic and real-world data. The synthetic subset consists of 3,000 images of size 512 × 512; their ground-truth (GT) counterparts are randomly cropped from DIV2K-Val\citep{agustsson2017ntire} and degraded via the RealESRGAN\citep{wang2021real} pipeline. The real-world evaluation data include LQ-HQ pairs from DRealSR\citep{wei2020component} and RealSR\citep{cai2019toward}, with image sizes of 128 × 128 and 512 × 512, respectively.

\subsubsection*{Compared methods} \label{subsubsec413}
We compare SCEESR with one-step DM-based SR methods OSEDiff\citep{yin2024one}, multi-step DM-based SR methods ResShift\citep{yue2023resshift}, StableSR\citep{wang2024exploiting}, SeeSR\citep{wu2024seesr} and PASD\citep{khan2023pasd} and GAN-based SR methods BSRGAN\citep{zhang2021designing}, RealESRGAN\citep{wang2021real} and FeMaSR\citep{chen2022real}. All comparative results are obtained using officially released codes or models. 

\subsubsection*{Evaluation Metrics} \label{subsubsec414}
To ensure a comprehensive and holistic evaluation of different methods, we employ a diverse set of full-reference and no-reference image quality assessment metrics. The full-reference metrics include: PSNR and SSIM, which assess pixel-wise fidelity to the ground-truth image. LPIPS\citep{zhang2018unreasonable} and DISTS\citep{ding2020image}, which measure perceptual similarity in a deep feature space. For distribution-level consistency, we use the Fréchet Inception Distance (FID)\citep{heusel2017gans} to quantify the similarity between the distributions of ground-truth and restored images. Additionally, we incorporate several no-reference metrics to evaluate perceptual quality without relying on ground-truth references, including NIQE\citep{mittal2012making}, MUSIQ\citep{ke2021musiq}, CLIPIQA\citep{wang2023exploring} and MANIQA\citep{yang2022maniqa}.

\subsection{Comparison with State-of-the-Arts} \label{subsec42}
\subsubsection*{Quantitative Comparisons} \label{subsubsec421}
\begin{table*}[t]
\centering
\caption{Quantitative comparison with state-of-the-art methods on both synthetic and real-world benchmarks. The best results of each metric are highlighted in red and second best results are highlighted in blue. ‘s’ denotes the number of diffusion reverse steps in the method.}
\label{tab:quantitative_comparisons}
\resizebox{\textwidth}{!}{
\begin{tabular}{@{}l|c|*{9}{c}@{}}
\toprule
\textbf{Datasets} & \textbf{Methods} & \textbf{PSNR$\uparrow$} & \textbf{SSIM$\uparrow$} & \textbf{LPIPS$\downarrow$} & \textbf{DIST$\downarrow$} & \textbf{FID$\downarrow$} & \textbf{NIQE$\downarrow$} & \textbf{MUSIQ$\uparrow$} & \textbf{CLIPIQA$\uparrow$} &  \textbf{MANIQA$\uparrow$} \\
\toprule
\multirow{9}{*}{DIV2K-Val} 
& BSRGAN & \textcolor{lightblue}{24.56} & \textcolor{lightblue}{0.6264} & 0.3356 & 0.2279 & 44.25 &	4.7519 & 61.15 & 0.5242 & 0.5065 \\
& RealESRGAN & 24.27 & \textcolor{lightred}{0.6366} & 0.3117 & 0.2146 & 37.67 & 4.6788 & 61.01 & 0.5271 & 0.5496 \\
& FeMaSR & 23.05 & 0.5883 & 0.3133 & 0.2064 & 35.89 & 4.7415 & 60.79 & 0.5993 & 0.5068 \\
& StableSR-s200 & 23.23 & 0.5718 & 0.3119 & 0.2053 & \textcolor{lightred}{24.48} & 4.7585 & 65.88 & 0.6764 & 0.6187 \\
& ResShift-s15 & \textcolor{lightred}{24.62} & 0.6173 & 0.3354 & 0.2218 & 36.15 & 6.8217 & 61.01 & 0.6065 & 0.5448 \\
& PASD-s20 & 23.11 & 0.5496 & 0.3578 & 0.2213 & 29.22 & \textcolor{lightred}{4.3621} & \textcolor{lightblue}{68.91} & 0.6782 & \textcolor{lightred}{0.6477} \\
& SeeSR-s50 & 23.65 & 0.6033 & 0.3203 & \textcolor{lightblue}{0.1974} & 25.94 & 4.8107 & 68.62 & \textcolor{lightred}{0.6931} & 0.6235 \\
& OSEDiff-s1 & 23.68 & 0.6098 & \textcolor{lightblue}{0.2948} & 0.1983 & 26.36 & 4.7101 & 67.93 & 0.6674 & 0.6142 \\
\rowcolor{lightpurple}
& SCEESR-s1 & 23.78 & 0.6082 & \textcolor{lightred}{0.2904} & \textcolor{lightred}{0.1959} & \textcolor{lightblue}{25.68} & \textcolor{lightblue}{4.6687} & \textcolor{lightred}{68.93} & \textcolor{lightblue}{0.6852} & \textcolor{lightblue}{0.6242} \\
\midrule
\multirow{9}{*}{DrealSR}
& BSRGAN & \textcolor{lightred}{28.73} & \textcolor{lightblue}{0.8025} & 0.2989 & 0.2168 & 155.69 & 6.5196 & 57.10 & 0.4911 & 0.4872 \\
& RealESRGAN & \textcolor{lightblue}{28.62} & \textcolor{lightblue}{0.8047} & \textcolor{lightblue}{0.2965} & \textcolor{lightred}{0.2095} & 147.67 & 6.6931 & 54.13 & 0.4415 & 0.4901 \\
& FeMaSR & 26.89 & 0.7568 & 0.3175 & 0.2242 & 157.83 & \textcolor{lightblue}{5.9077} & 53.67 & 0.5459 & 0.4413 \\
& StableSR-s200 & 28.01 & 0.7531 & 0.3288 & 0.2275 & 149.05 & 6.5245 & 58.47 & 0.6351 & 0.5593 \\
& ResShift-s15 & 28.43 & 0.7666 & 0.4013 & 0.2663 & 172.33 & 8.1256 & 50.54 & 0.5336 & 0.4575 \\
& PASD-s20 & 27.32 & 0.7064 & 0.3767 & 0.2538 & 156.19 & \textcolor{lightred}{5.5478} & 64.82 & 0.6801 & \textcolor{lightred}{0.6162} \\
& SeeSR-s50 & 28.14 & 0.7687 & 0.3196 & 0.2323 & 147.45 & 6.3978 & \textcolor{lightblue}{64.86} & 0.6798 & 0.6035 \\
& OSEDiff-s1 & 27.89 & 0.7829 & 0.2975 & 0.2172 & \textcolor{lightblue}{135.36} & 6.4912 & 64.58 & \textcolor{lightblue}{0.6956} & 0.5892 \\
\rowcolor{lightpurple}
& SCEESR-s1 & 28.13 & 0.7819 & \textcolor{lightred}{0.2963} & \textcolor{lightblue}{0.2167} & \textcolor{lightred}{133.93} & 6.3457 & \textcolor{lightred}{65.41} & \textcolor{lightred}{0.6965} & \textcolor{lightblue}{0.6054} \\
\midrule
\multirow{9}{*}{RealSR}
& BSRGAN & \textcolor{lightred}{26.37} & \textcolor{lightred}{0.7649} & \textcolor{lightblue}{0.2795} & 0.2125 & 141.36 & 5.6575 & 63.15 & 0.4997 & 0.5394 \\
& RealESRGAN & 25.67 & \textcolor{lightblue}{0.7611} & 0.2832 & \textcolor{lightblue}{0.2098} & 135.23 & 5.8299 & 60.12 & 0.4442 & 0.5482 \\
& FeMaSR & 25.06 & 0.7352 & 0.2948 & 0.2293 & 141.11 & 5.7891 & 58.90 & 0.5264 & 0.4861 \\
& StableSR-s200 & 24.67 & 0.7077 & 0.3025 & 0.2295 & 128.59 & 5.9131 & 65.71 & 0.6172 & 0.6215 \\
& ResShift-s15 & \textcolor{lightblue}{26.29} & 0.7416 & 0.3467 & 0.2507 & 141.77 & 7.2643 & 58.35 & 0.5437 & 0.5278 \\
& PASD-s20 & 25.18 & 0.6792 & 0.3388 & 0.2269 & 124.38 & \textcolor{lightblue}{5.4145} & 68.68 & 0.6613 & \textcolor{lightred}{0.6481} \\
& SeeSR-s50 & 25.14 & 0.7211 & 0.3015 & 0.2229 & 125.64 & \textcolor{lightred}{5.4092} & \textcolor{lightblue}{69.39} & 0.6604 & 0.6435 \\
& OSEDiff-s1 & 25.13 & 0.7336 & 0.2931 & 0.2135 & \textcolor{lightred}{123.56} & 5.6483 & 69.01 & \textcolor{lightblue}{0.6687} & 0.6318 \\
\rowcolor{lightpurple}
& SCEESR-s1 & 25.35 & 0.7386 & \textcolor{lightred}{0.2793} & \textcolor{lightred}{0.2091} & \textcolor{lightblue}{123.73} & 5.5861 & \textcolor{lightred}{69.41} & \textcolor{lightred}{0.6698} & \textcolor{lightblue}{0.6447} \\
\bottomrule
\end{tabular}
}
\end{table*}

Table \ref{tab:quantitative_comparisons} compares the performance of our proposed SCEESR-s1 model with various GAN-baed and DM-based approaches. The following findings can be made. Traditional methods like BSRGAN\citep{zhang2021designing}, RealESRGAN\citep{wang2021real} and FeMaSR\citep{chen2022real}, while achieving competitive PSNR and SSIM in some cases, often fall short in no-reference metrics such as FID\citep{heusel2017gans}, MUSIQ\citep{ke2021musiq}, and CLIPIQA\citep{wang2023exploring}. Among DM-based approaches, StableSR-s200\citep{wang2024exploiting}, despite achieving the best FID\citep{heusel2017gans} on DIV2K-Val, does not consistently lead in other perceptual metrics. ResShift-s15\citep{yue2023resshift} shows limited performance across both reference and no-reference metrics, often ranking lower than other DM methods. PASD-s20\citep{khan2023pasd} and SeeSR-s50\citep{wu2024seesr} achieves improved no-reference metric scores such as MUSIQ\citep{ke2021musiq}, CLIPIQA\citep{wang2023exploring}, and MANIQA\citep{yang2022maniqa}, but their multi-step inference makes them less efficient. Moreover, their relatively higher LPIPS\citep{zhang2018unreasonable} and DIST\citep{ding2020image} scores compared to SCEESR-s1 suggest a less faithful reconstruction to the ground truth. OSEDiff-s1\citep{yin2024one} efficiently distills multi-step generation into a single step, yielding strong LPIPS\citep{zhang2018unreasonable} and DIST\citep{ding2020image} scores. However, its no-reference metrics, while good, are often surpassed by methods like PASD\citep{khan2023pasd}, SeeSR\citep{wu2024seesr} and SCEESR-s1. Our proposed SCEESR-s1 distinguishes itself by maintaining high efficiency, while consistently achieving impressive perceptual quality. Notably, SCEESR-s1 excels not only in reference-based perceptual metrics such as LPIPS\citep{zhang2018unreasonable} and DIST\citep{ding2020image} but also consistently leads in no-reference metrics such as CLIPIQA\citep{wang2023exploring}, MUSIQ\citep{ke2021musiq}, and MANIQA\citep{yang2022maniqa}, demonstrating its superior ability to generate perceptually realistic and high-fidelity images with high efficiency across all evaluated datasets.

\subsubsection*{Qualitative comparisons} \label{subsubsec422}
\begin{figure}[htbp] 
    \centering
    \includegraphics[width=0.5\textwidth]{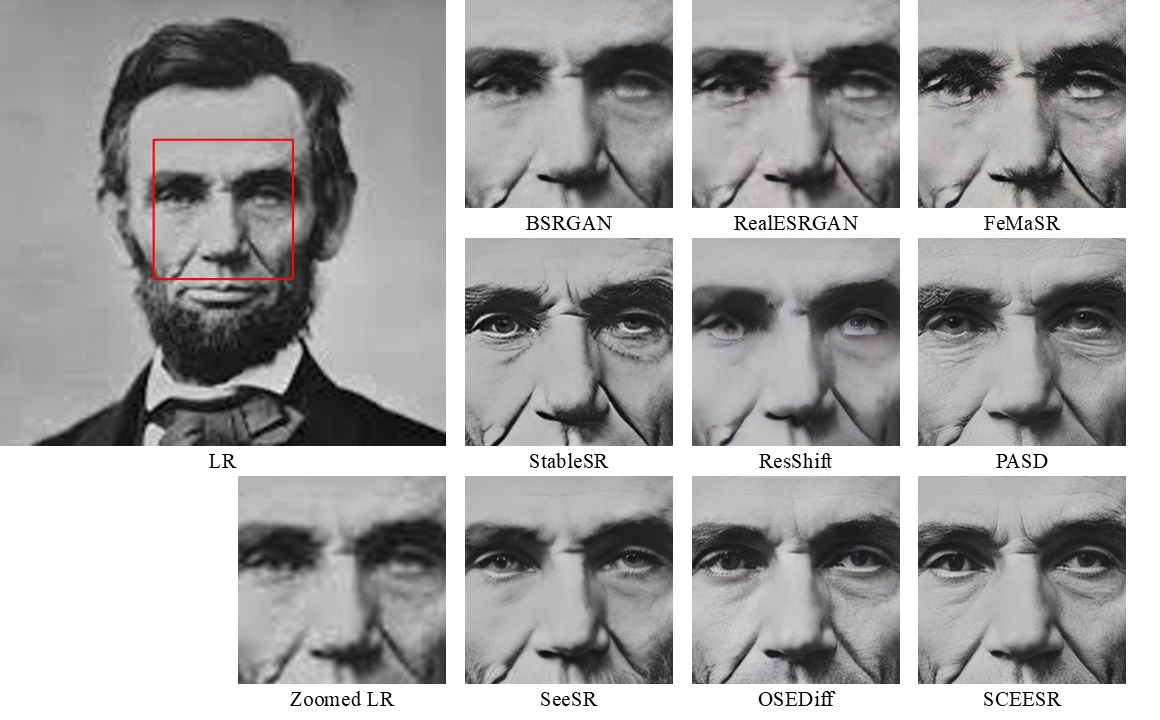}
    
    \caption{Vision comparisons of different GAN-based and DM-based SR methods on Lincoln's face.}
    
    \label{fig:compare_Lincoln}
\end{figure}

\begin{figure}[htbp] 
    \centering
    \includegraphics[width=0.5\textwidth]{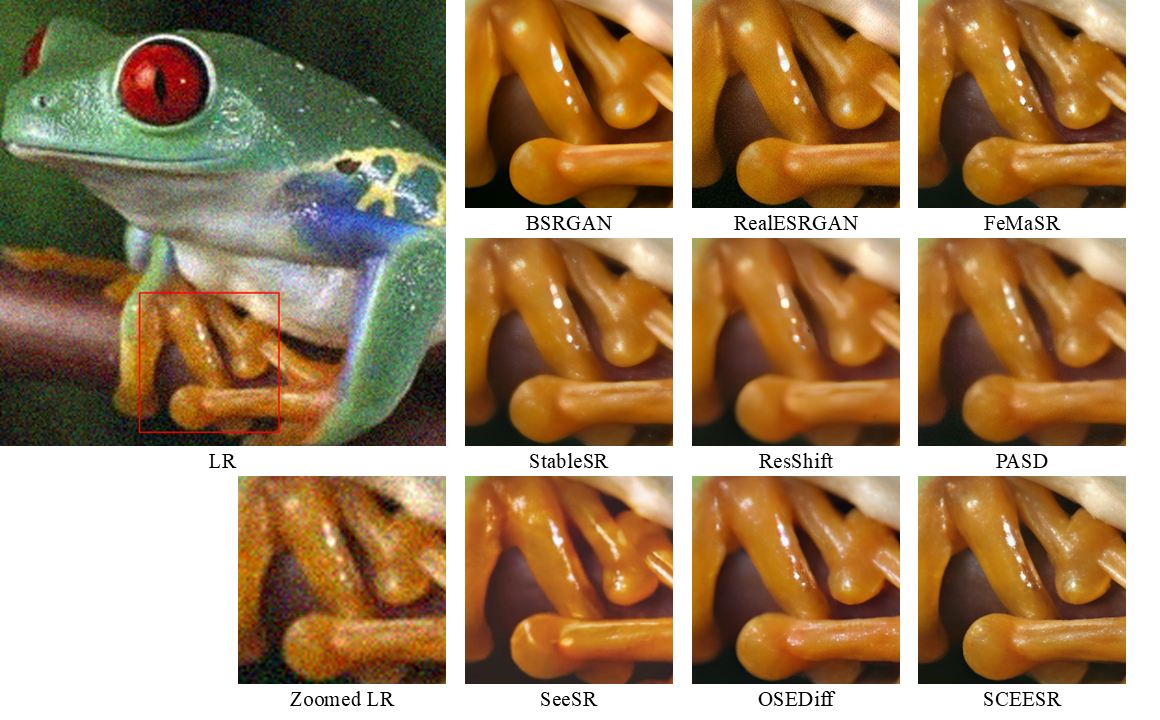}
    
    \caption{Vision comparisons of different GAN-based and DM-based SR methods on a frog's toes.}
    
    \label{fig:compare_frog}
\end{figure}

Qualitative comparisons of real-world image super-resolution methods are provided in Figure~\ref{fig:compare_Lincoln} and Figure~\ref{fig:compare_frog}. 

The first example, the face of Lincoln, reveals that the conventional GAN methods fail to reconstruct sharp facial details. Methods built upon Stable Diffusion (SD), such as ResShift\citep{yue2023resshift} and SeeSR\citep{wu2024seesr} blur the facial details, especially the wrinkles on the face. StableSR\citep{wang2024exploiting} and PASD\citep{khan2023pasd} often introduces overly exaggerated and unnatural textures. OSEDiff\citep{yin2024one} achieves realistic facial details. However, compared with OSEDiff\citep{yin2024one}, our SCEESR is more powerful in restoring edge information, such as wrinkles and boundaries. 

This superiority is further confirmed in the second example, the toes of a frog. BSRGAN\citep{zhang2021designing}, RealESRGAN\citep{wang2021real} and FeMaSR\citep{chen2022real} lack of color gradation and three-dimensionality due to their frameworks. StableSR\citep{wang2024exploiting}, lacking explicit semantic guidance and SeeSR\citep{wu2024seesr}, lacking extensive pre-trained image priors are constrained in generating rich and accurate textures. Although PASD\citep{khan2023pasd} incorporates text prompts, its non-robust prompt extraction under degradation leads to worse semantic generation. SeeSR\citep{wu2024seesr} attempts to address this with degradation-aware cues, but the resulting color appear unnatural, possibly due to instability from random noise sampling. OSEDiff\citep{yin2024one} succeeds in producing finely detailed and natural textures, but the edges of the toes are blurry. SCEESR ensures the rationality of the colors and enhances the clarity of the edges.

\subsubsection*{Complexity Comparisons} \label{subsubsec423}
\begin{table*}[t]
\centering
\caption{The inference time and the number of total parameters of DM-based SR methods.}
\label{tab:inference_comparisons}
\resizebox{0.6\textwidth}{!}{
\begin{tabular}{@{}l|*{6}{c}@{}}
\toprule
 & \textbf{StableSR} & \textbf{ResShift} & \textbf{PASD} & \textbf{SeeSR} & \textbf{OSEDiff} & \textbf{SCEESR} \\
\midrule
\multirow{1}{*}{Inference Steps} 
 & 200 & 15 & 20 & 20 & 1 & 1 \\
\midrule
\multirow{1}{*}{Inference Time(s)}
 & 11.40 & 0.72 & 2.70 & 4.30 & 0.12 & 0.15 \\
\midrule
\multirow{1}{*}{$\#$Total Parm(M)}
 & 1510 & 179 & 2210 & 2524 & 1775 & 2028 \\
\bottomrule
\end{tabular}
}
\end{table*}

A complexity comparison of diffusion-based SR models is presented in Table \ref{tab:inference_comparisons}, with metrics collected on an NVIDIA V100 GPU for 512×512 inputs. The results confirm the efficiency of the one-step inference paradigm. SCEESR maintains a significant speed advantage over multi-step models. When compared to the faster one-step model, OSEDiff\citep{yin2024one}, SCEESR introduces a modest increase in computational cost, which is a deliberate design trade-off that directly contributes to its notably higher reconstruction quality.

\subsection{Ablation Study} \label{subsec43}
\subsubsection*{Effectiveness of Controlnet and AME Loss} \label{subsubsec431}
\begin{table*}[t]
\centering
\caption{Quatitative comparison with state-of-the-art methods on both synthetic and read-world benchmarks between the results of the first part of SCEESR with only L2 loss and LPIPS\citep{zhang2018unreasonable} loss and the results of the second part of SCEESR with semantic control edge enhancement ControlNet and AME loss.}
\label{tab:effectiveness_comparisons}
\resizebox{\textwidth}{!}{
\begin{tabular}{@{}l|c|*{9}{c}@{}}
\toprule
\textbf{Datasets} & \textbf{SCEESR Phase} & \textbf{PSNR$\uparrow$} & \textbf{SSIM$\uparrow$} & \textbf{LPIPS$\downarrow$} & \textbf{DIST$\downarrow$} & \textbf{FID$\downarrow$} & \textbf{NIQE$\downarrow$} & \textbf{MUSIQ$\uparrow$} & \textbf{CLIPIQA$\uparrow$} &  \textbf{MANIQA$\uparrow$} \\
\toprule
\multirow{2}{*}{DIV2K-Val} 
& 1st & 23.63 & 0.6087 & 0.2925 & 0.1976 & 26.64 & 4.7162 & 68.46 & 0.6713 & 0.6135 \\
& 2nd & 23.78 & 0.6082 & 0.2904 & 0.1959 & 25.68 & 4.6687 & 68.93 & 0.6852 & 0.6242 \\
\toprule
\multirow{2}{*}{DrealSR}
& 1st & 27.83 & 0.7832 & 0.2971 & 0.2169 & 134.78 & 6.5139 & 63.24 & 0.6947 & 0.5886 \\
& 2nd & 28.13 & 0.7819 & 0.2963 & 0.2167 & 133.93 & 6.3457 & 65.41 & 0.6965 & 0.6054 \\
\toprule
\multirow{2}{*}{RealSR}
& 1st & 25.06 & 0.7354 & 0.2823 & 0.2114 & 123.68 & 5.6114 & 69.43 & 0.6696 & 0.6399 \\
& 2nd & 25.35 & 0.7386 & 0.2793 & 0.2091 & 123.73 & 5.5861 & 69.41 & 0.6698 & 0.6447 \\
\toprule
\end{tabular}
}
\end{table*}

\begin{figure}[htbp] 
    \centering
    \includegraphics[width=0.45\textwidth]{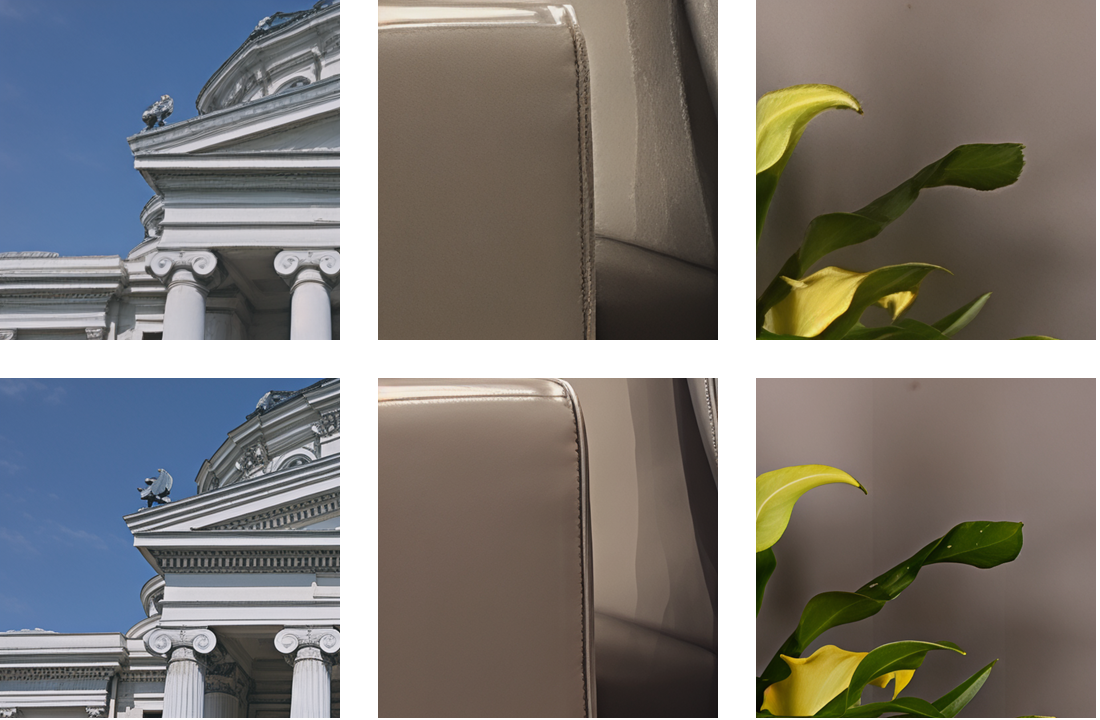}
    
    \caption{Qualitative comparison with state-of-the-art methods on both synthetic and read-world benchmarks between the results of the first phase of SCEESR with only L2 loss and LPIPS loss(the first row) and the results of the second phase of SCEESR with semantic control edge enhancement ControlNet and AME loss(the second row).}
    
    \label{fig:part_comparison__}
\end{figure}

To validate the effectiveness of our proposed Semantic Control Edge Enhancement ControlNet and Adaptive Multi-Detector Edge Loss (AME), we compare two training phases. The first phase uses only L2 and LPIPS\citep{zhang2018unreasonable} losses without ControlNet, while the second phase incorporates the AME loss along with our ControlNet module. Quantitative results presented in Table \ref{tab:effectiveness_comparisons} demonstrate that the key components of SCEESR play a crucial role in improving the quality of (SR images across multiple metrics. Furthermore, qualitative results in Figure~\ref{fig:part_comparison__} clearly show enhanced details and sharper edges in the reconstructed images. These findings confirm that the integration of the SCEE ControlNet and AME loss effectively boosts the perceptual and structural quality of SR outputs.

\section{Conclusion and Limitation} \label{sec5}
In this work, we introduce SCEESR, a novel super-resolution (SR) framework that enhances a pre-trained Stable Diffusion (SD) model via Low-Rank Adaptation (LoRA) modules and incorporates a semantic control edge enhancement module based on ControlNet. By leveraging edge-related information as input, SCEESR effectively improves both structural and perceptual quality. The framework employs the commonly used L2 loss to minimize pixel-wise regression errors and adopts the LPIPS\citep{zhang2018unreasonable} loss to reduce perceptual discrepancies. SCEESR demonstrates strong performance in terms of both effectiveness and efficiency. Moreover, with the proposed AME loss, it further enhances structural quality, thereby simultaneously improving perceptual performance.

Despite its competitive performance in enhancing structural and perceptual quality, SCEESR introduces a modest increase in inference time and memory overhead. Furthermore, extreme input images may lead to the generation of meaningless edge maps. In such cases, the erroneous information introduced by ControlNet could adversely affect the diffusion process. To address these limitations, future work will focus on developing more robust edge detectors and semantic control mechanisms with higher tolerance levels, thereby improving the reliability and effectiveness of the integrated ControlNet.

\section*{Acknowledgements}
We would like to express our sincere gratitude to Zhuang Yun Kai at ShanghaiTech University.

\section*{Statements and Declarations}
Competing Interests: The authors have no financial or proprietary interests in any material discussed in this article.

\noindent Funding: No funds, grants, or other support was received.

\noindent Data Availability: The datasets generated during and/or analyzed during the current study are available from the corresponding author on reasonable request.

\let\oldthebibliography\thebibliography
\renewcommand{\thebibliography}[1]{%
  \oldthebibliography{#1}%
  \footnotesize
}
\bibliography{sn-bibliography}% common bib file
%% if required, the content of .bbl file can be included here once bbl is generated
%%\input sn-article.bbl

\end{document}